\documentclass[11pt,a4paper]{article}
\usepackage{times,latexsym}
\usepackage{url}
\usepackage{graphicx}
\usepackage[T1]{fontenc}
\usepackage{booktabs}

\usepackage[acceptedWithA]{tacl2021v1}
\usepackage{xspace,mfirstuc,tabulary}
\usepackage{multirow}

\newcommand{\xsim}{\texttt{xsim}\xspace}
\newcommand{\xsimpp}{\texttt{xsim++}\xspace}
\newcommand{\flores}{\textsc{Flores}\xspace}
\newcommand{\fleurs}{\textsc{Fleurs}\xspace}

\newcommand{\laserthree}{\textsc{Laser3}\xspace}
\newcommand{\sonar}{\textsc{Sonar}\xspace}

\newcommand{\sonarurl}{\url{https://github.com/facebookresearch/SONAR}}

\newcommand{\NbLangsMined}{37\xspace} % P0, P1 and P2

\newif\iftaclinstructions
\taclinstructionsfalse % AUTHORS: do NOT set this to true
\iftaclinstructions
\renewcommand{\confidential}{}
\renewcommand{\anonsubtext}{(No author info supplied here, for consistency with
TACL-submission anonymization requirements)}
\newcommand{\instr}
\fi

\iftaclpubformat % this "if" is set by the choice of options

\else

\fi

\title{\sonar: Sentence-Level Multimodal \\and Language-Agnostic Representations}
\author{
  \hspace*{0pt} Paul-Ambroise Duquenne \\
  \hspace*{0pt} Meta AI \& Inria \\
  \hspace*{0pt} \texttt{\small padqn@meta.com} \\
  \And
  \hspace*{0pt} Holger Schwenk \\
  \hspace*{0pt} Meta AI \\
  \hspace*{0pt} \texttt{\small schwenk@meta.com}
  \And
  \hspace*{-25pt} Benoît Sagot \hspace*{-40pt}\\
  \hspace*{-25pt} Inria \hspace*{-40pt}\\
  \hspace*{-25pt} \texttt{\small benoit.sagot@inria.fr} \hspace*{-40pt} \\
}

\date{}

\begin{document}
\maketitle
\begin{abstract}
  
  We introduce \sonar, a new multilingual and multimodal fixed-size sentence embedding space. Our single text encoder, covering 200 languages, substantially outperforms existing sentence embeddings such as \laserthree and LabSE on the \xsim{} and \xsimpp{} multilingual similarity search tasks. Speech segments can be embedded in the same \sonar embedding space using language-specific speech encoders trained in a teacher-student setting on speech transcription data. Our encoders outperform existing speech encoders on similarity search tasks. %\pa{Maybe add 37 languages handled}
  We also provide a text decoder for 200 languages, which allows us to perform text-to-text and speech-to-text machine translation, including for zero-shot language and modality combinations. Our text-to-text results are competitive compared to the state-of-the-art NLLB~1B model, despite the fixed-size bottleneck representation. Our zero-shot speech-to-text translation results compare favorably with strong supervised baselines such as Whisper.
\end{abstract}

\section{Introduction}
Representation learning of sentences has been widely studied in recent years for different purposes: from classification of sentences \citep{devlin2018bert} to multilingual representations for translation purposes \citep{pham2019improving}.
Different pre-training objectives were explored to build contextual representations from sentences \citep{devlin2018bert,conneau2019unsupervised,clark2020electra}. However, these methods often lack sentence-level objectives, making it difficult to evaluate the semantic similarity between two sentences. On the other hand, several works focused on learning sentence embeddings \citep{cer2018universal,conneau2017supervised,reimers2019sentence}, aiming to encode sentences with similar meanings closely in the sentence embedding space.  \citet{Artetxe:2019:tacl_massive_ml,labse} extended this idea to multilingual sentences, enabling the semantic comparison between sentences from different languages. This was used to perform bitext mining at scale, to automatically align monolingual sentences from Common Crawl \citep{schwenk:2021:acl_ccmatrix}. 
This mined bitext data can be successfully used to train state-of-the art machine translation (MT) models \citep{schwenk:2021:acl_ccmatrix,costa2022no}.
In recent research, we may distinguish three main approaches to building multilingual fixed-size sentence representations. 
\begin{figure}[t!]
    \centering
    \includegraphics[width=1\linewidth]{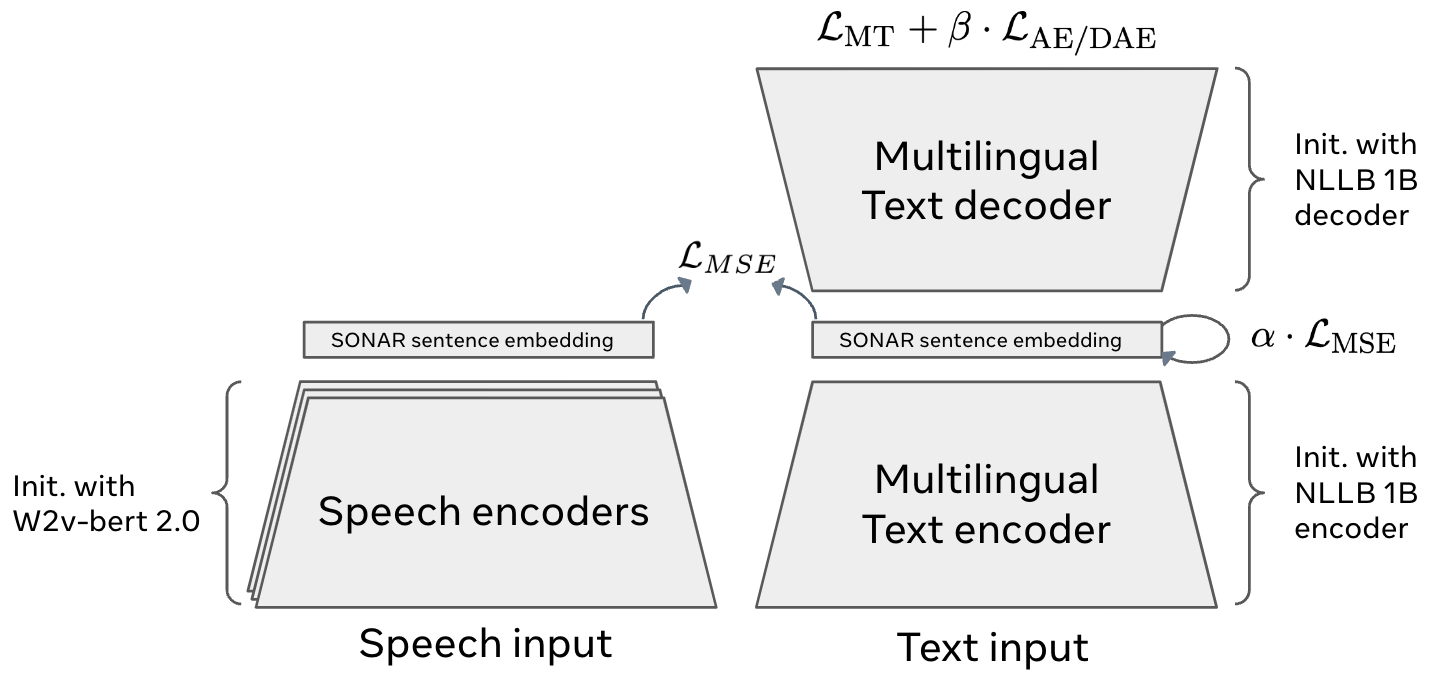}
    \caption{\sonar architecture.}
    \label{fig:architecture}
\end{figure}

\vspace*{-2mm}
\paragraph{Encoder-only approaches} such as \citep{labse}, which learn sentence embeddings for text, based on a siamese encoder architecture. Contrastive loss is often used to learn similar representations for different text translations while avoiding collapse (i.e. avoid to predict the same embedding for
every input)

\vspace*{-2mm}
\paragraph{Encoder-decoder approaches} such as \citep{Artetxe:2019:tacl_massive_ml}, which learn sentence embeddings with a translation objective, that can be computed thanks to an additional decoder. The main difference with classical sequence-to-sequence model is the bottleneck layer, or pooling function, that computes a fixed-size sentence representation between the encoder and the decoder. 

\paragraph{Teacher-student approaches} such as \citep{Reimers,laser3}, which extend a (possibly monolingual) pre-existing sentence embedding space to new languages with a teacher-student learning strategy. The existing embedding space is used as teacher to train student encoders for new languages. Bitext training data is used for this kind of training, where the sentence in the new language is encoded with a trained encoder, while its translation in another supported language is encoded with the pre-existing encoder as target. The same teacher-student approach can be used to extend a text-only multilingual sentence embedding space to the speech modality by training speech encoders \citep{Duquenne:2021:nips_mine,khurana2022samu}. These speech encoders can be used to perform speech-to-text or speech-to-speech translation mining \citep{duquenne2022speechmatrix}.

In this work, we used an encoder-decoder approach to build our sentence embedding space \sonar on text data only. We then used a teacher-student approach to train speech encoders for the same space.

Our motivation for using an encoder-decoder approach for the initial text-based training phase is twofold. First, a multilingual decoder is trained along the multilingual encoder, which opens possibilities such as zero-shot MT \cite{tmodules}. Second, a pre-trained state-of-the-art MT encoder-decoder model can be used to initialize the whole encoder-decoder architecture, in this work we used NLLB 1B dense model as initialization.
In contrast to previous work, we % explore a number of training strategies to build the sentence embedding space, studying %Holger: on dit deux fois la même chose dans une phrase
study the effect of different training objective functions on the properties of the resulting embedding space. More precisely, we combine translation, auto-encoding and denoising objectives, together with a cross-lingual similarity objective in the sentence embedding space. 

In a second step, we train speech student encoders using our multilingual text encoder as a teacher. We demonstrate the cross-modal similarity search and speech translation\footnote{The term ``speech translation'' customarily denotes speech-to-text translation.} capabilities of the resulting \sonar framework.

In summary, the main contributions of the \sonar (Sentence-level multimOdal and laNguage-Agnostic Representations) model are as follows:
\begin{itemize}
    \item We explore different training objectives to learn a multilingual sentence embedding space initialized from the NLLB 1B model, thoroughly comparing the different approaches on a wide range of decoding and similarity search evaluations
    \item This yield a single sentence encoder for 200 languages which significantly outperform state-of-the-art sentence embedding approaches;
    \item We trained speech encoders for \NbLangsMined languages using teacher-student training
    \item We provide a text decoder for 200 languages enabling (zero-shot) text and speech translation
    \item We analyzed the cross-lingual and cross-modal similarity search and decoding capabilities of our \sonar framework.
    \item The \sonar text and speech encoders as well as the text decoders are freely available at \sonarurl.
\end{itemize}

\section{Related work}

\paragraph{Multilingual sentence representations} Many works have studied how to efficiently learn multilingual representations of sentences. Some of them focused on variable-length representations of sentences, learning high-level contextual representations for each sub-word like multilingual BERT \citep{devlin2018bert} or XLM-R \citep{xlmr}. Others learnt fixed-size sentence representations by integrating sentence-level objectives in the training. It is the case for example of sentence-BERT \citep{reimers2019sentence}, which was initially trained on English text only, and later extended to other languages with a teacher-student approach \citep{Reimers}. The English model behaves as a teacher to train a multilingual encoder covering other languages. The student model is initialized with the XLM-R pretrained encoder and fine-tuned using bitext training data. The original English encoder, which is kept frozen, is used to generate an embedding for the English translation of each sentence, which then serves as a target for the student encoder via a regression loss.

Bitexts can also be used in other ways to train multilingual sentence embedding spaces. LASER \citep{Artetxe:2019:tacl_massive_ml} is an encoder-decoder architecture, with a fixed-size sentence representation
between the encoder and the decoder, trained with a translation objective. The orginal LASER covers 93 languages. Its decoder was originally used for training only, as the encoder itself defines the sentence embedding space. However, recent work such as \citep{tmodules} showed that it is possible to learn high quality decoders for LASER representations into multiple languages, thereby enabling zero-shot MT on unseen languages directions. Similarly to \citet{Reimers}, \citet{laser3} introduced \laserthree, extending LASER to new languages, including low-resource languages, using a teacher-student approach.
Finally, LaBSE \citep{labse} uses a
dual-encoder approach with and additive margin softmax objective \cite{yang2019improving}. It highlights the benefits of initializing encoders with multilingual pre-trained models and covers 109 languages.

\paragraph{Joint speech/text sentence representations}
There has been a large body or research on unsupervised representation learning for monolingual \citep{wav2vec2} and multilingual speech \cite{babu2021xls}, with recently w2v-bert \cite{chung2021w2v} that combines contrastive learning and masked language modeling to learn self-supervised representations from speech. Other works explored multilingual and multimodal (speech/text) pre-training methods, including mSLAM \citep{mslam}. Finally, \citet{Duquenne:2021:nips_mine}, followed by \citet{khurana2022samu}, introduced multilingual and multimodal sentence embeddings, extending a pre-existing multilingual text sentence embedding space to the speech modality with a distillation approach. \citet{tmodules,interspeech} also showed that it is possible to efficiently decode multilingual speech sentence embeddings with decoders trained on text sentence embeddings into different languages, to perform zero-shot speech translation.

\begin{figure*}
    \centering
    \small
    \begin{minipage}{1\textwidth}
        \textbf{English sentence from FLORES:}\\
        \textit{Dr. Ehud Ur, professor of medicine at Dalhousie University in Halifax, Nova Scotia and chair of the clinical and scientific division of the Canadian Diabetes Association cautioned that the research is still in its early days.} \\
        \textbf{Auto-encoding of the sentence with \sonar:}\\
        \textit{Dr. Ehud Ur, professor of medicine at Dalhousie University in Halifax, Nova Scotia and chairman of the clinical and scientific division of the Canadian Diabetes Association warned that the research is still in its early stages.}
    \end{minipage}
    \caption{Example of a long sentence with named entities auto-encoded with \sonar.}
\end{figure*}

\section{Methodology}
\label{sec:method}

To build our multilingual and multimodal sentence embedding space \sonar, we follow a two-step training strategy, inspired by \citet{Duquenne:2021:nips_mine,tmodules}. The first step is to build a sentence embedding space for text: we are building a multilingual sentence embedding space based on an encoder-decoder approach. The second step extends the multilingual text sentence embedding space to the speech modality, using a teacher-student approach.

\subsection{Multilingual sentence representations for text}

Contrarily to LASER's bidirectional LSTM architecture \citep{Artetxe:2019:tacl_massive_ml}, \sonar relies on a Transformer encoder-decoder architecture, initialized with pre-trained MT model weights. However, as opposed to standard sequence-to-sequence architectures for MT, the architecture we use to train \sonar on parallel text data goes through a single vector bottleneck that represents the full sentence and does not use token-level cross-attention. The fixed-size sentence representation is computed by pooling the token-level outputs of the encoder. Instead of doing cross-attention on a variable-length sequence of encoder outputs, the decoder only attends to this single vector at each decoding step. Different pooling methods can be used to compute this fixed-size representation, including max- and mean-pooling on token-level encoder outputs, as well as the encoder output for a special BOS token.

Contrarily to LASER \citep{Artetxe:2019:tacl_massive_ml}, we do not only train our encoder-decoder architecture using an MT objective only. We investigated several other objectives and combinations thereof and analyzed their effect on the sentence embedding space and the decoding performance of the resulting model. We introduce below the different objectives used to train our encoder-decoder architecture.

\paragraph{Translation objective} Following \citep{Artetxe:2019:tacl_massive_ml} work, we used parallel data to train our encoder-decoder architecture with a translation objective. To better understand the motivation behind this objective, let us take this example: Given a triplet of translations $x, y, z$, where $z$ is the English translation, decoding $x$ and $y$ into English may be easily achieved by the decoder if the sentence representation of these two input sentences are similar in the sentence embedding space. Training a encoder-decoder architecture on a translation objective may end up in this potential local minimum where translations are encoded closely to one another, so as to be decoded into the same target language sentence. However, there is no guarantee to converge to this local minimum. Nothing explicitly constrains a sentence in a language and its translation in another language to be encoded closely to one another. As a result, other local minima are possible, where translations are not encoded closely but still decoded into the same sentence for a given target language. To mitigate this, shallow decoders were used by \citet{Artetxe:2019:tacl_massive_ml}: a deeper decoder can more easily decode different points into the same sentence, whereas a shallower decoder is more  likely to need two vectors to be very similar whenever they must be decoded into the same sentence.

\paragraph{Auto-encoding and denoising auto-encoding objective} Auto-encoders have been widely used to build representations. It has the advantage to encourage encoding fine-grained details of the input. However, this objective by itself is not likely to learn semantic representation of sentences. Moreover, this objective is much simpler to learn compared to a translation objective, which makes the combination of these two objectives difficult. To mitigate these issues, \citet{liu2020multilingual} introduce a \textit{denoising} auto-encoding task, which has proven to be a good pre-training objective for translation tasks.

\paragraph{MSE loss objective in the sentence embedding space} Teacher-student approaches to  multilingual sentence embedding space learning have shown that ensuring that translations of a same sentence are embedded close to one another in the sentence embedding space with an MSE loss works really well \citep{Reimers,laser3}. However, using this kind of loss without a frozen pre-existing teacher embedding space would lead to collapse (all inputs mapped to the same embedding), which is why contrastive learning methods were introduced to learn multilingual sentence embeddings from scratch \citep{labse}. However, combining an MSE loss with a translation objective and/or a denoising auto-encoding objective could also prevent collapse from happening, as the model is forced to keep embeddings distinct to encode and decode different sentences.

\paragraph{Decoder finetuning}
\citet{tmodules} demonstrated that learning deep decoders for an existing sentence embedding space (in their case, LASER) can significantly improve translation and auto-encoding performance. While keeping the existing embedding space unchanged, such decoders greatly improve the decoding of sentence embeddings, therefore significantly improving auto-encoding and translation performance when combined with compatible encoders. This is of great interest for zero-shot (possibly cross-modal) translation, as shown by \citet{interspeech}.

In this paper, we introduce a decoder fine-tuning method called \textit{random interpolation decoding}. Based on a trained encoder-decoder model with a bottleneck representation between the encoder and the decoder, we freeze the encoder weights and fine-tune the decoder weights only on a specific decoding task: Given a bitext $x, y$, we encode $x$ and $y$ with the frozen encoder,  randomly draw $z$ as a random interpolation of $x$ and $y$ embeddings, and learn to decode sentence embedding $z$ into $y$. This can be viewed as a continuous combination of translation and auto-encoding tasks.

\subsection{Multilingual sentence representations for speech}
\citet{Duquenne:2021:nips_mine} introduced the first semantic sentence embeddings for multilingual speech. Their method follows a teacher-student approach, where the teacher model is an encoder for multilingual sentence embeddings trained on text. We follow the same approach but using our newly trained text sentence embedding space as teacher: we trained a speech student encoder to encode audios into fixed-size representations and minimize the MSE loss between the transcription sentence embeddings and the trained speech sentence embeddings. Written translation embeddings could also be used as targets in this teacher-student approach \citep{Duquenne:2021:nips_mine}. However, in this work, we only focus on transcriptions as targets, using written translations is left for future work. As in previous work, we leveraged self-supervised pre-trained models, for our speech encoders training, using a w2v-bert pretrained model as initialization.

\begin{table*}[t]
\centering\small
\begin{tabular}{llll|ll}
\toprule
Method & X-eng$\uparrow$ &  eng-X$\uparrow$  & AE$\uparrow$ & \xsim{}$\downarrow$ & \xsimpp{}$\downarrow$ \\
\midrule
$\mathcal{L}_{\mathrm{MT}}$ & 33.2 & 21.1 & 28.6 & 1.3 & 19.6\\
$\mathcal{L}_{\mathrm{MT}} + \mathcal{L}_{\mathrm{AE}}$ & 17.6 & 18.6 & 94.6 & 15.9 & 65.7 \\
$\mathcal{L}_{\mathrm{MT}} + 0.1 \cdot \mathcal{L}_{\mathrm{DAE}}$ & 31.6 & 20.9 & 41.6 & 2.6 & 26.2 \\
$\mathcal{L}_{\mathrm{MT}} + 0.1 \cdot \mathcal{L}_{\mathrm{MSE}}$ & 31.7 & 20.2 & 27.2 & 1.3 & 14.3 \\
  \hline
 \multicolumn{1}{|l}{\textbf{\sonar sentence embedding space}}  &  & & & & \multicolumn{1}{l|}{} \\

\multicolumn{1}{|l}{$\mathcal{L}_{\mathrm{MT}} + 0.1 \cdot \mathcal{L}_{\mathrm{MSE}} + 0.01 \cdot 
 \mathcal{L}_{\mathrm{DAE}}$} & 32.9 & 20.7 & 32.4  & 1.4 & \multicolumn{1}{l|}{15.2}\\
 \multicolumn{1}{|l}{$\mathcal{L}_{\mathrm{MT}} + 0.1 \cdot \mathcal{L}_{\mathrm{MSE}} + 0.01 \cdot 
 \mathcal{L}_{\mathrm{DAE}}$ \& fine-tuned dec.} &  32.7 & 21.6 & 41.7 & 1.4 &  \multicolumn{1}{l|}{15.2} \\
\hline
\midrule
\textbf{MT topline}  &  & & & & \\
NLLB 1B & 35.2 & 24.9 & $39.0^{*}$ & $3.7^{*}$ & $49.6^{*}$ \\
\textbf{Similarity search baselines}  &  & & & & \\
LaBSE  & --- & --- & ---  & 10.7 & 36.1 \\
\laserthree  & --- & --- & ---  & 5.1 & 36.4 \\
 \bottomrule
\end{tabular}
\caption{Text evaluations on FLORES200 devtest set, averaged on the 200 languages supported by NLLB 1B: translation spBLEU for X-eng and eng-X directions, auto-encoding spBLEU, \xsim{} and \xsimpp{} similarity search results on X-eng pairs. Results with * are zero-shot evaluations of NLLB 1B model which was not trained to optimize these tasks.}
\label{tab:t2t_eval}
\end{table*}

\section{Evaluations}
To evaluate the semantic properties of the resulting sentence embedding space, we relied on a number of evaluation tasks on both text and speech modalities:

\subsection{Evaluations on text}

\paragraph{\xsim}
Cross-lingual similarity search, also called \xsim{},\footnote{\url{https://github.com/facebookresearch/LASER}} evaluates the similarity between sentence embeddings across languages. Given a test dataset of bitexts, translations are encoded into the multilingual sentence embedding space and cosine similarity between all embeddings are computed. For each test instance, if the two corresponding translations are not the closest, we count it as an error in order to compute an error rate on the whole test set.

\paragraph{\xsimpp}
More recently, \xsimpp{} was introduced as a more semantically challenging similarity search task \citep{chen2023xsim++}.\footnotemark[2] It augments the test set with hard negative examples for the similarity search, generating several modified versions of ground truth examples based on causality alternation, entity replacement and number replacement.

\paragraph{Translation tasks}
Multilingual embeddings are decoded into other target languages to perform MT. We report spBLEU (flores200) scores and COMET scores on the generated translations. Decoding sentence embeddings into other languages partially evaluates how much information is encoded in sentence embeddings, which is complementary to \xsim{} and \xsimpp{} evaluations. However, please note that information may also be restored from the internal language modeling capabilities of the decoder, and not from the sentence embeddings themselves.

\paragraph{Auto-encoding task}
Similarly to translation tasks, we decode sentence embedding in the same language to perform auto-encoding and evaluate the content preservation of this operation.

All these evaluations for text were performed on FLORES-200 devtest set,\footnote{\url{https://github.com/facebookresearch/flores/tree/main/flores200}} which provides an $N$-way parallel corpus of translations in 200 languages.

\subsection{Evaluations on speech}

\paragraph{\xsim{} for speech}
We follow \citet{Duquenne:2021:nips_mine} and calculate cross-modal and -lingual similarity search on the \fleurs{} speech translation test set \cite{conneau2023fleurs}. It follows the \xsim{} evaluation presented above, but \xsim{} is run on speech embeddings against English text translation embeddings.

\paragraph{\xsimpp{} for speech} In addition to \xsim{} computation for speech, we augment the English texts with challenging negative examples from the \xsimpp{} modified English sentences of FLORES.

\paragraph{Zero-shot speech-to-text translation}
Following \citet{tmodules}, speech student encoders can be combined with text decoders at inference time. Since the speech encoder were trained on ASR data only and the \sonar text decoder was only trained on text and has never seen speech embeddings during training, this corresponds to zero-shot speech-to-text translation. Similarly to text, it enables evaluating the content encoding in the speech embeddings. It also evaluates the compatibility between speech and text representations.
    
\paragraph{Zero-shot Automatic Speech Recognition:}
we also decode speech embeddings in the same language to perform speech recognition.

All these evaluations for speech were performed on \fleurs{} test set \citep{conneau2023fleurs}, a \mbox{$N$-way} parallel speech dataset in 102 languages built on top of the text \flores{}-101 benchmark.

\section{Experiments on text}
\label{experiments_text}
In this paper, we first trained a multilingual sentence embedding space using an encoder-decoder architecture on text, with fixed-representation of sentences between the encoder and the decoder. 
\subsection{Training setup}
We initialized our model with the NLLB 1B dense model \citep{costa2022no}, that was trained on translation tasks with full cross-attention on variable length encoder outputs as it is commonly done for sequence-to-sequence MT model training. The model is composed of a 24 layers Transformer encoder and a 24 layers Transformer decoder and trained on a combination of human labeled data, back-translated data and mined data \citep{costa2022no}. In order to build our fixed-size sentence representation, we added a pooling operation on the encoder outputs. Several pooling methods are possible: max-pooling as done in \citep{Artetxe:2019:tacl_massive_ml}, mean-pooling as done in \citep{reimers2019sentence}, or EOS pooling which use the output representation of the EOS special token appended at the end of sentences during NLLB training. Contrary to mean-pooling or EOS-pooling, max-pooling outputs a vector with a different range of values compared to NLLB training (due to the max operation), leading to worse results in our initial experiments. Since for EOS-pooling the training happened to be unstable during initial experiments, we focused on mean-pooling for the rest of our experiments. We trained our encoder-decoder model for 100k updates with same learning rate and batch size as NLLB training in the following experiments, unless explicitly specified. We used all bitext data used in the NLLB training, human labeled bitexts, back-translated data and mined data. This training dataset involves 200 target languages which contrasts with LASER training that only used English and Spanish as target languages. As presented in Section~\ref{sec:method}, we ran an extensive study on the use of different training objectives, namely translation objective (MT), auto-encoding objective (AE), denoising auto-encoding objective (DAE) and Mean Squared Error loss (MSE) in the sentence embedding space:
\[
\mathcal{L} = \mathcal{L}_{\mathrm{MT}} + \alpha \cdot \mathcal{L}_{\mathrm{MSE}} + \beta \cdot 
 \mathcal{L}_{\mathrm{AE/DAE}}
\]
We are using the same training data for auto-encoding and translation objectives, inputting the target sentences instead of the source sentences to perform auto-encoding of target sentences only. Incorporating more monolingual data in the training for the auto-encoding task is left to future work.

\subsection{Initial experiment with translation objective only}
We report the results of our experiments on text sentence embedding modeling in Table~\ref{tab:t2t_eval}. Our first experiment using only the translation objective for our encoder-decoder model with fixed-size intermediate representation gives surprisingly good translation performance, given the bottleneck between the encoder and the decoder. It yields -2 BLEU on X-eng direction and -3.8 BLEU on eng-X direction compared to NLLB 1B model with full-cross attention. 

We notice that auto-encoding evaluation (AE) significantly lags behind NLLB 1B model. This result may come from an inductive bias of the sequence-to-sequence architecture with full cross-attention, that could bias the model towards copying encoder inputs. 

\xsim{} and \xsimpp{} results are significantly better compared to previous work, namely LaBSE and \laserthree, on our 200 languages of focus, with approximately 45\% relative reduction of \xsimpp{} error rate compared to the baseline models. Note that averaging NLLB 1B encoder outputs to perform similarity search already gives good \xsim{} scores. This directly comes from the translation objective used during NLLB 1B training that encourages to encode multilingual sentences in similar ways for efficient cross-lingual transfer. However, the more difficult \xsimpp{} evaluation remains challenging, in this zero-shot setting, for the original NLLB 1B model.

\subsection{Experiments with auto-encoding objectives}
Noticing the gap in the auto-encoding performance between the fixed-size bottleneck encoder-decoder model and NLLB 1B, we integrated an auto-encoding objective, hoping to close the gap with the NLLB 1B model. This model was only trained for 50k steps, as it converged quickly compared to other variants. We notice that auto-encoding task is easy to learn, even with a fixed-size bottleneck between the encoder and the decoder, almost reaching 95 BLEU in average on the 200 languages of NLLB. This shows that a lot of information can be efficiently stored in a fixed-size representation and that the bottleneck should not be seen as an hard limitation. While the translation performance on eng-X tranlation directions is not that much impacted, we see a big drop in translation performance for X-eng directions (-15,6 BLEU) compared to the fixed-size bottleneck encoder-decoder model trained only on a translation task. Moreover, we see a big drop in both \xsim{} and \xsimpp{} evaluations showing that the model is not learning language-agnostic representations anymore, due to the auto-encoding objective that seems more easily optimized compared to the translation objective.

To mitigate the negative effects of the auto-encoding objective, while improving the auto-encoding performance at inference time, we switched to a denoising auto-encoding criterion, to avoid that the model overfits on the \textit{copy} task. That would also make the task harder compared to simple auto-encoding and could be better combined with the non-trivial translation task. We also scaled down this denoising auto-encoding objective by a factor 0.1. This mostly mitigated the performance drops on translation tasks, while significantly boosting the auto-encoding task (+13 BLEU) compared to the translation-only model. However, the denoising auto-encoding criterion significantly affects the \xsim{} and \xsimpp{} scores. This again shows that auto-encoding affects the organization of the sentence embedding space, learning distinct representations for different languages to optimize auto-encoding.

\begin{table}[t]
\centering\small
\begin{tabular}{lll}
\toprule
Method & X-eng &  eng-X \\
\midrule
\sonar & 85.9 &  83.4 \\
 \sonar \& fine-tuned dec. & 85.9 &  84.2 \\
\midrule
\textbf{Topline}  &  & \\
NLLB 1B & 86.5 & 85.2\\
 \bottomrule
\end{tabular}
\caption{Translation evaluations for X-eng and eng-X directions on FLORES200 devtest set: COMET scores averaged on 89 languages supported by both COMET and NLLB 1B models.}
\label{tab:comet_eval}
\end{table}

\subsection{Experiments with cross-lingual similarity objective}
Motivated by the recent distillation approaches to extend a sentence embedding space to new languages, explicitly aligning languages with an MSE criterion in the embedding space, we explored the use of an auxiliary MSE loss in the sentence embedding space. This is in addition to the translation loss, with the hope to explicitly make translations closer in the embedding space. In Table~\ref{tab:t2t_eval}, we notice that this new constraint degrades the decoding performance of the encoder-decoder model for both translation and auto-encoding tasks. However, it significantly boosts the \xsimpp{} scores compared to the encoder-decoder model trained only on a translation task, with -5.3 \xsimpp{} error rate reduction. 
\begin{table}[!t]
\centering\small
\resizebox{0.48\textwidth}{!}{
\begin{tabular}{lrrrr}\toprule
                               & fra  & spa  & swh  & rus  \\
                               \midrule
\textbf{X-eng BLEU}                     &      &      &      &      \\
\sonar \& fine-tuned dec.          & 46.1 & 34.5 & 42.4 & 37.1 \\
\laserthree\textsubscript{MSE} \& T-mod. &  40.4   &  29.6    &   27.2  &  29.7    \\
\midrule
\textbf{\xsimpp{}}                         &      &      &      &      \\
\sonar          & 4.8  & 7.9  & 7.1  & 6.5  \\
\laserthree\textsubscript{MSE}                    &   7.6   &  12.6   &   15.2   &  12.4   \\
\bottomrule
\end{tabular}}
\caption{
%\todo{update with final results}
Comparison to T-modules framework based on LASER embedding space. spBLEU scores for X-eng translation directions on FLORES200 devtest set and \xsimpp{} for X-eng pairs on FLORES200 devtest set.}
\label{tab:tmod_comparison}
\end{table}

\subsection{Training the \sonar embedding space}
Based on the conclusions of the previously trained models, we combined the translation loss, the auxiliary MSE loss and the denoising auto-encoding loss, to create the \sonar embedding space. In this run, the denoising auto-encoding loss is further downscaled, motivated by the high \xsimpp{} score of the previously trained sentence embedding space trained on denoising auto-encoding. First, in the same tendency from previous training with (denoising) auto-encoding objective, we notice a slight degradation in \xsimpp{} scores when adding the denoising auto-encoding in addition to the MSE loss. However, this degradation is only 0.9\% which can be considered as acceptable. Initial experiments on larger scaling factors for the denoising auto-encoding criterion further increased, as expected, the \xsimpp{} degradation, and we thus decided to stick with a 0.01 scaling factor for the denoising auto-encoding objective. On the other hand, for our new \sonar model, we see improvements on translation tasks compared to the model trained on MT and MSE loss. This may be due to efficient mitigation of collapse that could happen with MSE loss, thanks to the denoising auto-encoding objective.  We also see big improvements in auto-encoding task (>+3.8 BLEU) compared to all models not trained with auto-encoding objectives. This variant seems to be the best setup in terms of sentence embedding space organization (following \xsim{} and \xsimpp{} scores) and decoding performance (following translation and auto-encoding evaluations). We also report the \xsim{} and \xsimpp{} results on the intersection of languages handled by LaBSE, LASER3 and \sonar in Table~\ref{tab:sonar_xsim}, and notice again that \sonar outperforms previous state-of-the-art sentence embedding spaces for multilingual similarity search.

\begin{table}[!t]
    \centering
    \begin{tabular}{lcc}
        \toprule
        & \multicolumn{2}{c}{98 languages} \\
        &  \xsim$\downarrow$  &\xsimpp$\downarrow$ \\
        \midrule
        \sonar  & 0.1 & 9.3 \\
        \laserthree  & 1.1 & 27.5 \\
        LaBSE & 1.5 & 15.4 \\
        \bottomrule
    \end{tabular}
    \caption{Comparison of similarity search results (error rates) on the intersection of languages handled by LaBSE,
LASER3 and \sonar.}
    \label{tab:sonar_xsim}
\end{table}

Finally, we tried to improve the decoding performances of our architecture, freezing the embedding space and our multilingual encoder, while fine-tuning only the decoder. We used the \textit{random interpolation decoding} method introduced in section \ref{sec:method}, where we compute a random interpolation of the source and target sentence embeddings and learn to decode the target sentence tokens. As the encoder is frozen, the \xsim{} and \xsimpp{} scores won't change, but the decoding results will. With this decoder fine-tuning step, we notice similar translation results on the X-eng direction, while noticing a +0.9BLEU gain on the eng-X translation directions. More importantly, the auto-encoding performance is boosted by 9.3 BLEU with decoder fine-tuning method while the sentence embedding space was not affected. This finetuning step is trained for 50k additional steps.

We also evaluated the best performing models on translation tasks with COMET, which has proven to better correlate with human judgments compared to BLEU scores. We evaluated the two X-eng and eng-X directions involving the languages on which XLM-R was trained on, which are the languages supported by COMET (see Table~\ref{tab:comet_eval}). We see less that 1 point difference between our \sonar encoder-decoder model (with fine-tuned decoder) compared to NLLB 1B model for both eng-X and X-eng directions, showing the good quality of the translations.

The NLLB 1B model still represents a topline, and to evaluate our \sonar framework against a more fair baseline involving a fixed-size sentence representation between the encoder and the decoder, we compared our results to the decoding of LASER embeddings, recently introduced in T-modules \citep{tmodules,interspeech}.  As \laserthree encoders were trained with a cosine loss, the sentence embeddings cannot be efficiently decoded with T-modules decoder. This is why we trained new \laserthree encoders with MSE loss, and added back-translated data from NLLB project in addition to the original training data of \laserthree encoders. These newly trained \laserthree\textsubscript{MSE} encoders can be combined with T-modules decoder \citep{interspeech} to perform X-eng translation. We report the results on 4 languages French, Spanish, Swahili and Russian in Table~\ref{tab:tmod_comparison} and notice big improvements using \sonar on both X-eng translation task and \xsimpp{} evaluation . Please note that compared to previous work \citep{tmodules}, we are able to encode and decode 200 languages with a single encoder and a single decoder.

\begin{table}[!t]
\centering\small
\begin{tabular}{lll}
\toprule
BLEU       & fra-eng  & spa-eng \\
\midrule
\sonar mean-pooling & 25.2 & 20.6 \\
\sonar max-pooling  & 31.6 & 24.5 \\
\sonar attention-pooling   & 33.3 & 25.5 \\
\bottomrule
\end{tabular}
\caption{spBLEU X-eng zero-shot speech translation on \fleurs{} test set for different pooling methods.}
\label{tab:pooling_ablation}
\end{table}

\begin{table}[!b]
\centering\small
\begin{tabular}{lllll}
\toprule
                      & \textbf{fra} & \textbf{spa}  & \textbf{swh} & \textbf{rus} \\
\midrule
\textbf{xsim} & & & & \\
\sonar    & 0.0         & 0.0         & 0.0         & 0.0         \\
\laserthree\textsubscript{MSE} & 0.0         & 0.0         & 0.0        & 0.3         \\
\midrule
\textbf{\xsimpp{}} & & & & \\
\sonar   & 12.3       & 13.9       & 22.8       & 24.6        \\
\laserthree\textsubscript{MSE} & 17.5       & 24.9        & 40.7       & 42.1   \\
\bottomrule
\end{tabular}
\caption{Multilingual and multimodal similarity search evaluations on \fleurs{} test set: \xsim{} and \xsimpp{} error rates on speech translation X-eng pairs.}
\label{tab:s2t_xsim}
\end{table}

%=================================================
\section{Experiments on speech}
\label{experiments_speech}
Based on the experiments and evaluations of multilingual sentence embedding spaces for text, we chose to focus only on the embedding space learnt with translation, denoising auto-encoding and MSE objectives which seems to be a good trade-off between good semantic representation (\xsim{} and \xsimpp{}) and good decoding performance (translation and auto-encoding). We follow a teacher-student approach to extend this space to the speech modality for several languages. We first performed an initial extensive study on five languages only: English (eng), Spanish (spa), French (fra), Russian (rus) and Swahili (swh). We then scale to \NbLangsMined languages.

\begin{table}[!t]
\centering\small
\resizebox{0.48\textwidth}{!}{
\begin{tabular}{@{}lllll@{}}
\toprule
                                  & \textbf{fra}   & \textbf{spa}   & \textbf{swh}   & \textbf{rus}   
\\
\midrule
\textbf{Training hours}          &               &               &               &               \\
\sonar/LASER ASR  &      0.8k         &      0.4k         & 0.3k          & 0.2k         \\
Whisper ASR  &      10k         &      11k         & 0.01k          & 10k         \\
Whisper ST  &      4k         &      7k         & 0.3k          & 8k         \\

\hline

\textbf{\sonar zero-shot ST}          &               &               &               &               \\
\sonar                & 33.3 & 25.5 & 14.9          & 15.0          \\
\sonar \& fine-tuned dec. & 33.4          & 24.8          & 15.6          & 14.6 \\
\midrule
\textbf{Zero-shot ST baseline}                                         &               &               &               &               \\
\laserthree\textsubscript{MSE} \& T-mod            & 30.7          & 22.9          & 3.7 & 16.2           \\
\midrule
\textbf{Supervised ST toplines}                                         &               &               &               &               \\
Whisper Large v1                  & 33.8          & 27.0          & 5.2          & 30.2           \\
Whisper Large v2             & 34.9          & 27.2          & 7.6          & 31.1       \\   
\bottomrule
\end{tabular}
}
\caption{spBLEU scores on \fleurs{} test set for zero-shot Speech Translation on X-eng directions.}
\label{tab:s2t_eval}
\end{table}

\begin{table}[!b]
\centering\small
\begin{tabular}{@{}lllllll@{}}
\toprule
src\textbackslash{}tgt & \textbf{eng} & \textbf{fra} & \textbf{spa} & \textbf{swh} & \textbf{rus} & \textbf{\begin{tabular}[t]{c} 200 \\ langs \end{tabular}}\\
\midrule
eng & 69.7 & 44.3 & 26.9 & 27.8 & 29.8  & 17.7\\
fra & 33.4 & 64.1 & 21.5 & 18.2 & 23.3 & 13.4\\
spa & 24.8 & 25.1 & 58.9 & 16.0 & 16.8 & 11.7\\
swh & 15.6 & 13.5 & 9.0  & 25.7 & 9.8  & 7.0\\
rus & 14.6 & 17.3 & 11.0 & 10.4 & 35.0  & 8.0\\
\bottomrule
\end{tabular}
\caption{spBLEU scores on \fleurs{} test set for zero-shot Speech Translation on \{eng,fra,spa,swh,rus\}-X directions. Last column is the average spBLEU Speech Translation scores for decoding in the 200 languages supported by \sonar text decoder.}
\label{tab:s2t_200_eval}
\end{table}

\begin{table*}[!t]
\centering\small
\begin{tabular}{llllll}
\toprule
                       & \textbf{eng}   & \textbf{fra}   & \textbf{spa}   & \textbf{swh}   & \textbf{rus}   \\
\midrule
\textbf{Training hours}          &               &               &               &               \\
\sonar/LASER ASR  & 4k &      0.8k         &      0.4k         & 0.3k          & 0.2k         \\
Whisper ASR  &   438k  & 10k         &      11k         & 0.01k          & 10k         \\
Whisper ST  &    ---  & 4k         &      7k         & 0.3k          & 8k         \\
\midrule
\textbf{BLEU}                   &       &       &       &       &       \\
\sonar                  & 64.7  & 54.3  & 50.0  & 17.7  & 29.1  \\
\sonar \& fine-tuned dec  & 69.7  & 64.1  & 58.9  & 25.7  & 35.0  \\
\midrule
Whisper v1             & 80.8  & 79.8  & 84.8  & 26.9  & 84.3  \\
Whisper v2             & 81.3  & 82.0  & 85.3  & 36.0  & 85.3  \\
\midrule
\textbf{Bert-score}             &       &       &       &       &       \\
\sonar                  & 0.948 & 0.926 & 0.923 & 0.808 & 0.853 \\
\sonar \& fine-tuned dec  & 0.951 & 0.939 & 0.936 & 0.831 & 0.870 \\
\midrule
Whisper v1             & 0.972 & 0.965 & 0.977 & 0.837 & 0.975 \\
Whisper v2             & 0.972 & 0.969 & 0.979 & 0.865 & 0.978 \\
\bottomrule
\end{tabular}
\caption{Speech recognition spBLEU scores and Bert-scores on \fleurs{} test set.}
\label{tab:ars_eval}
\end{table*}

\subsection{Experiments on 5 languages}
\label{5speechlangs}
We use a pre-trained w2v-bert 600 million parameter model to initialize the speech encoders and train them on Common Voice 12 ASR training set \citep{ardila2019common}. For our English speech encoder, we also used ASR training data from Must-C \citep{mustc:2019:naacl}, Voxpopuli \citep{wang-etal-2021-voxpopuli} and Librispeech \citep{panayotov2015librispeech}. We tested different pooling methods, namely mean-pooling, max-pooling and attention-pooling. Attention-pooling is performed with a three layer transformer decoder architecture with cross-attention on the speech encoder outputs, in order to output a single vector as our speech sentence embedding. Best results are achieved with attention-pooling (see Table~\ref{tab:pooling_ablation}). %(See appendix for more details).

As a baseline, we compared our \sonar speech encoders to speech encoders trained with LASER as teacher (using our newly trained \laserthree\textsubscript{MSE} text encoders as teacher), with exact same training data and pre-trained w2v-bert model. We report the \xsim{} and \xsimpp{} cross-lingual and cross-modal results in Table~\ref{tab:s2t_xsim} on \fleurs{} test set for foreign speech embeddings against English text embeddings. Similarly to what \citet{chen2023xsim++} noticed on FLORES, \xsim{} scores saturate to zero error rate on \fleurs{} test set, not providing useful insights on the multimodal sentence embedding space organization. Therefore, we also report \xsimpp{} scores, augmenting the \fleurs{} test set with hard negatives generated in \citep{chen2023xsim++} based on FLORES which composes the source sentences of \fleurs{}. We notice 41\% \xsimpp{} relative reduction when switching from LASER as teacher to \sonar as teacher.

Following \citep{tmodules}, we decoded the speech sentence embeddings with our \sonar text decoder, performing zero-shot speech-to-text translation. Indeed, the text decoder has never seen speech sentence embeddings during training. Moreover, speech representations were only trained to match their transcription representations but never translations. In Table~\ref{tab:s2t_eval}, we report our zero-shot speech-to-text translation results on \fleurs{} test set for X-eng directions and compare it to the baseline trained on LASER space. We also report the state-of-the-art results for speech-to-text translation, trained in a supervised way on significantly more training data. First, we notice big improvements in the BLEU scores compared to the LASER baseline on French, Spanish an Swahili, with an average 5.5 BLEU gain on these languages, while being slightly behind on Russian to English translation (-1.2 BLEU). This last result is surprising, as our \sonar speech encoder have much better \xsimpp{} score on Russian compared to the LASER speech encoder. Second, we notice that for our two high resource languages, namely French and Spanish, our zero-shot speech-to-text results are close to Whisper Large v1 supervised results, while being trained on much less training data. As for Swahili, our framework significantly outperforms Whisper models. We notice much better results for Russian-to-English for Whisper which was expected given the amount of training data and the supervised setting.

Thanks to the compatibility across modalities and across languages, we decoded English, French, Spanish, Swahili and Russian speech sentence embeddings into the 200 text languages supported by our \sonar decoders. We report the zero-shot speech translation results using the fine-tuned \sonar decoder in Table~\ref{tab:s2t_200_eval}. We notice that BLEU scores remain high for other languages than English, still in a zero-shot setting, highlighting again the compatibility between representations.

Finally, speech embeddings can be decoded into text in the same language, which can be seen as speech transcription. Since our model can often paraphrase transcriptions, we report in Table~\ref{tab:ars_eval} BLEU scores as well as bert-scores for this zero-shot transcription task. While being significantly behind on BLEU scores, which is expected as our model often paraphrases transcriptions, we see much less gap with Whisper transcriptions with the bert-score metric (which still advantages real transcriptions compared to paraphrases, but less than BLEU). Training data amount is also significantly different between the two setups, but it's interesting to notice that the gap in terms of bert-score remains reasonable. 

%---------------------------------------

\subsection{Scaling to \NbLangsMined languages}

We use the same recipe than described above to extend the coverage of the speech encoders to \NbLangsMined languages. These speech encoders were trained by linguistic language family, e.g. Romance or Indian languages, using speech transcriptions only, from public and licensed sources. Table~\ref{tab:bleu37} column "Train" gives statistics on the amount of training data.
As in Section \ref{5speechlangs}, we evaluate the speech encoders by connecting them to the \sonar text decoder and calculate speech-to-text translation performance, as measured by BLEU. % BLEU13 or spmBLEU
Although our results are fully zero-shot speech translation,
we achieve very competitive performance compared to the state-of-the-art model Whisper v2 large \citep{radford2022robust}.
The average on BLEU scores are slightly better for SONAR compared to Whisper, while being zero-shot speech translation. Our model performs less well on some high-resource languages like Mandarin Chinese, German or French, but outperforms Whisper for others like Spanish or Dutch and for several less common languages, like Swahili or Uzbek.
Our modular approach seems to achieve particular good results on Indian languages: Bengali, Hindi, Kannada, Telugu, Tamil and Urdu.

\begin{table}[!t]
    \small
    
\begin{tabular}[t]{@{}ll|r|rr@{}} \\
\toprule
%\multirow{2}{*}{\bf ISO} & \multirow{2}{*}{\bf Language}  & \bf Train & \bf Decode & \bf Whisper  \\
% & & \textrm{audio [h]}  & \textrm{BLEU13} & \bf BLEU13a  \\
 ISO & Language & Train & Ours & Whisper \\
\midrule
\textrm{arb}  & \textrm{MS Arabic} & {822} & {30.9} & {26.9}\\
\textrm{ben}  & \textrm{Bengali} & {335} & {21.3} & {14.1}\\
\textrm{cat}  & \textrm{Catalan} & {1738} & {37.7} & {36.9}\\
\textrm{cmn}  & \textrm{Mandarin Chinese} & {9320} & {18.6} & {20.8}\\
\textrm{ces}  & \textrm{Czech} & {181} & {32.0} & {30.3}\\
\textrm{cym}  & \textrm{Welsh} & {99} & {14.5} & {13.4}\\
\textrm{dan}  & \textrm{Danish} & {115} & {34.9} & {36.0}\\
\textrm{deu}  & \textrm{German} & {3329} & {36.2} & {38.8}\\
\textrm{est}  & \textrm{Estonian} & {131} & {26.1} & {21.2}\\
\textrm{fin}  & \textrm{Finish} & {184} & {24.9} & {25.2}\\
\textrm{fra}  & \textrm{French} & {2057} & {33.7} & {34.9}\\
\textrm{hin}  & \textrm{Hindi} & {150} & {22.6} & {24.2}\\
\textrm{ind}  & \textrm{Indonesian} & {269} & {28.7} & {31.9}\\
\textrm{ita}  & \textrm{Italian} & {588} & {29.3} & {27.5}\\
\textrm{jpn}  & \textrm{Japanese} & {17319} & {20.2} & {20.8}\\
\textrm{kan}  & \textrm{Kannada} & {114} & {21.4} & {13.1}\\
\textrm{kor}  & \textrm{Korean} & {316} & {17.1} & {24.2}\\
\textrm{mlt}  & \textrm{Maltese} & {106} & {24.4} & {16.2}\\
\textrm{nld}  & \textrm{Dutch} & {1723} & {29.3} & {28.4}\\
\textrm{pes}  & \textrm{Western Persian} & {386} & {24.4} & {20.9}\\
\textrm{por}  & \textrm{Portuguese} & {269} & {38.3} & {41.4}\\
\textrm{pol}  & \textrm{Polish} & {304} & {21.1} & {25.8}\\
\textrm{ron}  & \textrm{Romanian} & {135} & {34.7} & {34.1}\\
\textrm{rus}  & \textrm{Russian} & {259} & {28.4} & {31.1}\\
\textrm{slk}  & \textrm{Slovak} & {102} & {32.3} & {29.3}\\
\textrm{spa}  & \textrm{Spanish} & {1511} & {28.0} & {27.2}\\
%\textrm{swe}  & \textrm{Swedish} & {144} & {} & {}\\
%\textrm{slv}  & \textrm{Slovenian} & {65} & {26.3} & {20.0}\\
\textrm{swh}  & \textrm{Swahili} & {361} & {23.5} & { 7.6}\\
\textrm{tam}  & \textrm{Tamil} & {245} & {16.2} & { 10.0}\\
\textrm{tel}  & \textrm{Telugu} & {84} & {18.0} & {14.7}\\
\textrm{tgl}  & \textrm{Tagalog} & {108} & {14.6} & {26.8}\\
\textrm{tha}  & \textrm{Thai} & {195} & {16.9} & {17.8}\\
\textrm{tur}  & \textrm{Turkish} & {174} & {22.7} & {29.9}\\
\textrm{ukr}  & \textrm{Ukrainian} & {105} & {30.7} & {32.5}\\
\textrm{urd}  & \textrm{Urdu} & {185} & {19.7} & {18.1}\\
\textrm{uzn}  & \textrm{Uzbek} & {115} & {20.0} & { 6.6}\\
\textrm{vie}  & \textrm{Vietnamese} & {194} & {19.1} & {21.9}\\
\midrule
\multicolumn{2}{l}{\textrm{Total/avg}} & {43628} & {25.3} & {24.5}\\
\bottomrule
\end{tabular}

    \caption{spBLEU evaluation of S2T into English on \fleurs{} test set. Our models were trained on ASR transcriptions only, compared to the Whisper large v2.% which uses a large amount of end-to-end S2T training data.
    }
    \label{tab:bleu37}
\end{table}

%---------------------------------------
\section{Discussion}
From all the experiments present in Section \ref{experiments_text} and Section \ref{experiments_speech}, we can draw a couple of high-level conclusions:

First, we have seen that the auto-encoding task can be greatly solved even with a fixed-size bottleneck between the encoder an the decoder, showing that a fixed-size representation should not be seen as a hard limitation, as a lot of information can be stored in a single vector.
Then, similarly to \citet{Artetxe:2019:tacl_massive_ml}, we noticed that a translation objective is well suited to build language-agnostic representations while making sure that the encoder model encodes enough information in the sentence embedding to be efficiently decoded (in another language). Adding an MSE loss in the training, which explicitly encourages to align languages in the sentence embedding space, leads to better language-agnostic representations. Moreover, denoising auto-encoding combined with MSE loss, can bring gains for decoding tasks, but too much of it affects the language-agnostic representations. Finally, teacher-student approach to extend to the speech modality has once again proven to be effective and the mutual compatibility between speech and text multilingual embeddings is greatly highlighted by the fact that speech embeddings can be decoded in foreign text in a zero-shot way.

\section{Conclusion}
To conclude, we introduced a new multilingual and multimodal sentence embedding space called \sonar. We conducted an extensive study on objective functions to build our multilingual teacher sentence embedding space for text, and an extensive evaluation of our \sonar framework for both similarity search and decoding tasks. We extended this new text sentence embedding space to the speech modality to introduce Sentence-level multimOdal and laNguage-Agnostic Representations (\sonar). The \sonar text and speech encoders as well as the text decoders are freely available at \sonarurl.

\section{Acknowledgment}
We would like to thank Kevin Heffernan for his help on providing \xsim{} and \xsimpp{} baselines for LaBSE and \laserthree, Andy Chung for providing the w2v-bert pre-trained models used as initialization for our speech encoders, Changhan Wang for providing speech data manifests used for training and Artyom Kozhevnikov and Naji El Hachem for the migration of models to fairseq2 for open-sourcing.

The last author's contribution was partly funded by his chair in the PRAIRIE institute funded by the French national agency ANR as part of the ``Investissements d'avenir'' programme under the reference ANR-19-P3IA-0001.

\newpage
\bibliography{tacl2021}
\bibliographystyle{acl_natbib}
\end{document}

\newpage

\iftaclpubformat

\onecolumn

\appendix
\section{Author/Affiliation Options as set forth by MIT Press}
\label{sec:authorformatting}

Option 1. Author’s address is underneath each name, centered.

\begin{quote}\centering
  \begin{tabular}{c}
    \textbf{First Author} \\
    First Affiliation \\
    First Address 1 \\
    First Address 2 \\
    \texttt{first.email@example.com}
  \end{tabular}
  \ 
  \begin{tabular}{c}
    \textbf{Second Author} \\
    Second Affiliation \\
    Second Address 1 \\
    Second Address 2 \\
    \texttt{second.email@example.com}
  \end{tabular}

  \begin{tabular}{c}
    \textbf{Third Author} \\
    Third Affiliation \\
    Third Address 1 \\
    Third Address 2 \\
    \texttt{third.email@example.com}
  \end{tabular}
\end{quote}

Option 2. Author’s address is linked with superscript characters to its name,
author names are grouped, centered.

\begin{quote}\centering
    \textbf{First Author$^\diamond$} \quad \textbf{Second Author$^\dagger$} \quad
    \textbf{Third Author$^\ddagger$}
    \\ \ \\
    $^\diamond$First Affiliation \\
    First Address 1 \\
    First Address 2 \\
    \texttt{first.email@example.com}
     \\ \ \\
     $^\dagger$Second Affiliation \\
    Second Address 1 \\
    Second Address 2 \\
    \texttt{second.email@example.com}
     \\ \ \\
    $^\ddagger$Third Affiliation \\
    Third Address 1 \\
    Third Address 2 \\
    \texttt{third.email@example.com}
\end{quote}
  
\fi

\end{document}